\documentclass{IEEEcsmag}

\usepackage[colorlinks,urlcolor=blue,linkcolor=blue,citecolor=blue]{hyperref}
\usepackage{cite}
\usepackage{multirow}
\usepackage{amsmath,amssymb,amsfonts}
\usepackage{algorithmic}
\usepackage{graphicx}
\usepackage{textcomp}
\usepackage{breqn}
\usepackage{wrapfig}
\usepackage{upmath}

\jvol{XX}
\jnum{XX}
\paper{8}
\jmonth{May/June}
\pubyear{2019}

\begin{document}


\title{RARITYNet: Rarity Guided Affective Emotion Learning Framework}

\author{Monu Verma}
\affil{University of Miami, 33155 USA}

\author{Santosh Kumar Vipparthi}
\affil{Indian Institute of Technology, Guwahati, 781039 India}


\begin{abstract}
Inspired from the assets of handcrafted and deep learning approaches, we proposed a RARITYNet: RARITY guided affective emotion learning framework to learn the
appearance features and identify the emotion class of facial expressions. The RARITYNet
framework is designed by combining the shallow (RARITY) and deep (AffEmoNet) features to
recognize the facial expressions from challenging images as spontaneous expressions,
pose variations, ethnicity changes, and illumination conditions. The RARITY is proposed to encode the inter-radial transitional patterns in the local neighbourhood.  The AffEmoNet: affective emotion learning network is proposed by incorporating three feature streams: high boost edge filtering (HBSEF) stream, to extract the edge information of highly affected facial
expressive regions, multi-scale sophisticated edge cumulative (MSSEC) stream is to learns the sophisticated edge information from multi-receptive fields and RARITY uplift complementary context feature (RUCCF) stream refines the RARITY-encoded features and aid the MSSEC stream features to enrich the learning ability of RARITYNet.

\end{abstract}

\maketitle

\chapterinitial{The introduction} Emotion is characterized by the mental state, feelings and intent of a human being. Facial expressions provide the salient information about the emotional state of a person. Automated human emotion recognition has many applications in medical diagnosis, depression analysis, behavioral profiling, gaming, human-computer interaction (HCI), lie detection, surveillance, marketing, e-learning system, etc. With the advancement in computing technologies and availability of large datasets, human emotion recognition has emerged as one of the important fields of research. Moreover, in order to develop computationally intelligent facial expression recognition (FER) system for real-world applications, it is important to evaluate FER techniques over a wide variety of challenging scenarios.  \par
 Feature extraction is one of the most important task for developing a robust FER system. Inadequate extraction of facial features would lead to performance degradation even with best of the classifier. Therefore, it is imperative to design robust feature descriptors to represent the facial expressions.
Feature extraction approaches can be characterized into two categories: predesigned and learning based. Predesigned (descriptor based) feature extraction approaches extract the features by designing handcrafted filters. Furthermore, these resultant features are used in the classification step to label the expression classes. However, in learning-based approaches, features are automatically learned from the training data using neural network architectures. The descriptor especially appearance-based features characterize the appearance of the entire facial area or selected region of an expression image \cite{shan2009facial, jabid2010robust, rivera2015local,ryu2017local}. The LBP and its variants \cite{shan2009facial} are sensitive to noise and are unable to handle intra-class variations. Similarly, the descriptors \cite{jabid2010robust, rivera2015local, ryu2017local} which exclusively utilize the gradient magnitudes, generate unstable patterns in smooth regions. Thus, it degrades the performance of the FER system in challenging visual conditions.\par
The existing handcrafted feature descriptors are focusing on the inter-class variations only. However, to define the disparities between emotion classes both inter and intra-class variations play decisive  roles. Therefore, to take the benefit of inter and intra-class  variations in this paper, we designed a novel descriptor: RARITY. RARITY utilizes the inter radial transitional information between the rings at multiple radius. This leads to identification of the edge spreads in the local  neighborhood. These edge spreads capture the edges using higher distanced neighbors and therefore are more robust to intra-class variations among facial expressions in an emotion class. Since, the RARITY patterns are encoded using pixel triplet relationship between two rings of different radius. This enhances the discriminability by highlighting the salient features (maximal and minimal disparities) and suppressing the noise elements in an image. Similarly, the bidirectional information improves the illumination invariance capability of the proposed descriptor. \par
Although Deep Learning architectures have shown impressive performance over the facial descriptors. These architectures have the ability to learn the appropriate features and classify the expression images accordingly.  Khorrami et al. \cite{khorrami2015deep} introduced a three-layer convolution neural network (CNN) to learn AU based features. Furthermore, Li et al. \cite{li2018eac} designed an enhancing and cropping (EAC) network using the detected landmarks to learn the salient features in each AU region. Lopes et al. \cite{lopes2017facial} created a big dataset by applying preprocessing steps on images to enhance the robustness of CNN network. Ding et al. \cite{ding2017facenet2expnet} designed a two-stage network: in the first stage, the convolutional layers are trained for the specific regions of an expression using the Face2exp net architecture and then, in the second stage, the fully-connected layers are concatenated to classify the expressions. Moreover, Yang et al. \cite{yang2020cascaded}, proposed a CNN based framework to makes full use of
extracted features by cascaded pyramids. They also replace
the backward propagation optimization to alleviate the redundant features. Similarly, other CNN based FER techniques \cite{mollahosseini2016going,lee2020multi, liu2021point} have also been proposed in the literature. \par
The Deep Learning networks (CNNs) require high computational and experimental cost.  This makes finding the appropriate technique highly increasing the overall computation and time cost of the CNNs. Moreover, analyzing as well as visualizing of features in an intuitive fashion is a pretty hard task for Deep Learning over handcrafted descriptors. Understanding the actual model with handcrafted features most of the time is more straightforward and less ambiguous. Facial expression recognition systems need an effective and efficient algorithm for real life challenges: spontaneous expressions, pose variations, ethnicity changes, and illumination conditions etc. with real time response on platforms having limited computational resources.  In this paper, we propose a framework: RARITYNet that bridge the gap between effective but inefficient methods and efficient but less effective methods.
\begin{figure*}[!t]
\centering
\centerline{\includegraphics[width=\linewidth,height=2.8in]{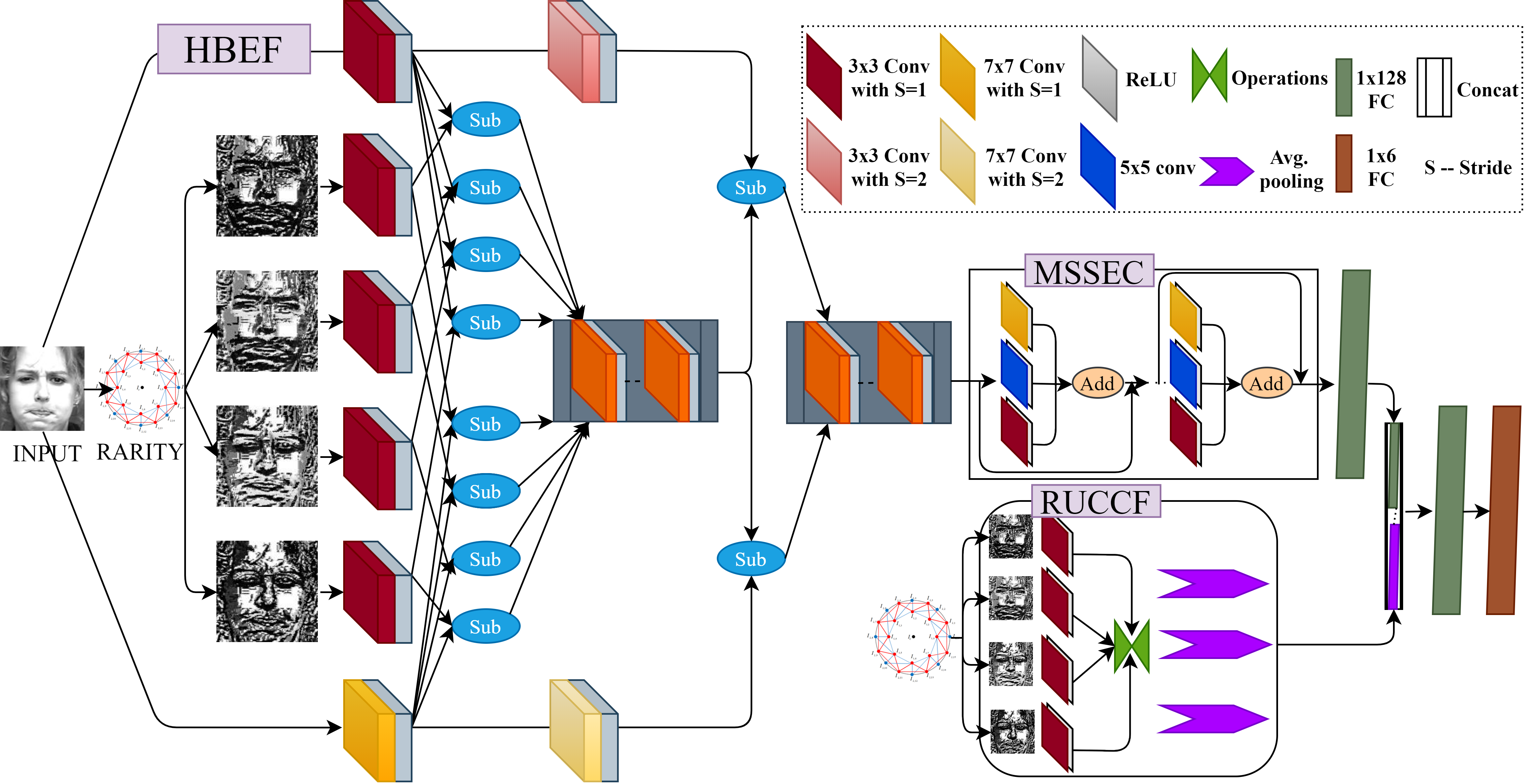}}
\caption{The proposed RARITYNet framework.}
\label{fig1}
\end{figure*}

\section{Proposed Method}
In this paper, we have proposed a RARITY guided affective emotion learning framework to learn the appearance features and identify the emotion class of facial expressions in various challenges: multi-view angles, candid, ethnicity variations, low resolution, blurriness, tiny faces, and lighting changes, captured by cameras sensors in various environments. The proposed framework contains two main blocks: RARITY and AffEmoNet to capture shallow and deep features, respectively as shown in Fig. \ref{fig1}. 

\subsection{RARITY}
\begin{figure}[!t]
\centering
\centerline{\includegraphics[width=0.9\columnwidth,height=2.4in]{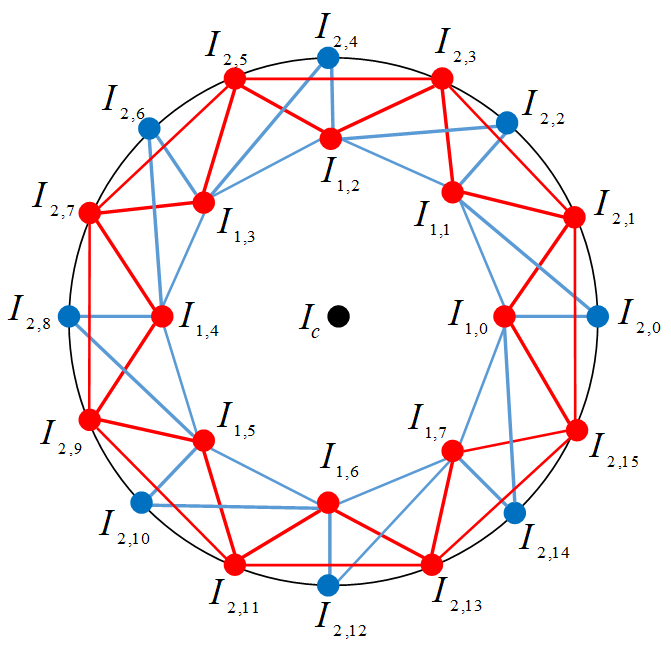}}
\caption{The proposed RARITY descriptor.}
\label{fig2}
\end{figure}

We propose a new feature descriptor named inter radial ring topology (RARITY) for facial expression recognition. Inspired by the phenomenon of ring topology, the RARITY is designed to encode the inter radial transitional patterns in an image to represent the texture variations in the local region. These transitional patterns are generated by establishing pixel triplet relations. The pixel triplets are selected from multiple radius in the local neighborhood. The inter-radial patterns, generated based on pixel triplet relations, assist in extracting the maximal and minimal disparities between the multi-radius ring topologies. Furthermore, we identify the bidirectional transition information between higher to lower and lower to higher radius at a given location. These patterns reveal the bright to dark and dark to bright variations in the local region. We illustrate the RARITY descriptor in Fig. \ref{fig2}. Moreover, the proposed descriptor captures significant edge information, which can effectively distinguish between the facial expressions.\par
The properties of RARITY are as follows: (1) The inter radial feature encoding enables the robust detection of coarse level edge spread to address the intra-class variations, (2) The pixel triplet relationship cohesively describes the inter radial variations. This property conceals the trivial information and highlights the salient expression changes, (3) The bi-directional relations between the rings at multiple radii extracts the first order gradient information. This increases the proposed descriptor’s robustness to noise and illumination variations.To compute the RARITY patterns, three inter radial pixels are selected to calculate the response code. For example, at $00$ direction ($I_1$,$0$, $I_2$,$0$), the pixels connected through the blue and red lines represent the two inter radial pixel sets as shown in Fig. \ref{fig2}. The triplet relation is encoded in bidirectional manner to extract the inter-radial transitional patterns. Thus, the RARITY generates four encoded response for a single image. Let $I(a,b)$ be a gray-scale image of size $M\times N$ where $a\in [1, M]$, $b\in [1,N]$. If at each location $I_c$ in the image, $p$ and $q$ neighborhood pixels are situated at radius $r_1$ and $r_2$. Then the RARITY (R) is computed using [Eq. \ref{eq1} - Eq. \ref{eq8}].
\begin{equation}\label{eq1}
    R(I_c)=\left \{\sum_{i=0}^{p-1}J_{\left \lceil \eta/2 \right \rceil}\left ( T\left ( r_k,i \right ) ,  U\left ( r_k,i \right ) \right )\times 2^i \right \}_{\eta=1}^{4}
\end{equation}
\begin{equation}\label{eq2}
   k=mod\left(\eta, 2\right)+1
\end{equation}
\begin{equation}\label{eq3}
   T\left(r_k, i\right)=I_{(r_k), i\times k}+I_{f_1, f_2}
\end{equation}
\begin{equation}\label{eq4}
   U\left(r_k, i\right)=I_{(r_k), i\times k}+I_{f_1, f_3}
\end{equation}
For each inter-radial pixel set, the RARITY encodes two patterns (bi-directional relations). Thus, for two inter-radial pixel sets, $\eta=4$  patterns are calculated in Eq. \ref{eq1}. The functions $f_1,f_2,f_3$ are defined through [Eq. \ref{eq5} - Eq. \ref{eq8}].
\begin{equation}\label{eq5}
  f_1=r_{k-(-1)^k}
\end{equation}
\begin{equation}\label{eq6}
    f_2=mod\left ( \psi \left ( k \right )\ast +(-1)^k, \psi \left ( k \right )\ast p \right )
\end{equation}
\begin{equation}\label{eq7}
  f_3=\psi \left(k\right)\ast i+mod\left(k,2\right)
\end{equation}
\begin{equation}\label{eq8}
  \psi\left(k\right)=\left \lfloor k/2 \right \rfloor+2\times mod\left(k,2\right)
\end{equation}
The functions $J_1\left(\cdot , \cdot \right)$  and $J_2\left(\cdot , \cdot \right)$ are computed using Eq. \ref{eq9} and Eq. \ref{eq10}.
\begin{equation}\label{eq9}
  J_1\left(x,y\right)=\left\{\begin{matrix}
1 & x>0\&\&y>0\\ 
0 & otherwise
\end{matrix}\right.
\end{equation}
\begin{equation}\label{eq10}
  J_2\left(x,y\right)=\left\{\begin{matrix}
1 & x<0\&\&y<0\\ 
0 & otherwise
\end{matrix}\right.
\end{equation}
   
\subsection{AffEmoNet}
The AffEmoNet  contains three feature streams: high boost edge filtering (HBSEF) stream, multi-scale sophisticated edge cumulative (MSSEC) stream and RARITY uplift complementary context feature (RUCCF) stream as shown in Fig. \ref{fig1}
\begin{table*}[!t]
\centering
\caption{Performance Comparison on CK+, MUG and ISED Datasets.}
\label{tab1}
\begin{tabular}{lccccccccc}
\hline
\multirow{2}{*}{\textbf{Method}} & \multirow{2}{*}{\textbf{Type}} & \multicolumn{2}{c}{\textbf{CK+}} & \multicolumn{2}{c}{\textbf{MUG}} & \multicolumn{2}{c}{\textbf{ISED}}&{\textbf{\#Parameters}}&{\textbf{\#Memory }}\\ \cline{3-8}
                        &                       & 6-EXP         & 7-EXP       & 6-EXP        & 7-EXP       & 4-EXP         & 5-EXP & (in millions) & (in mega bytes)    \\ \hline \hline
LBP \cite{shan2009facial} & HC & 89.97  & 83.96& 82.65  & 76.16& 63.29 & 59.00&NA&NA \\
LDP \cite{jabid2010robust} & HC& 90.84 & 84.80 & 82.87 & 78.70& 64.42  & 59.04&NA&NA \\
LDN \cite{rivera2012local}  & HC& 88.84 & 83.80 & 81.96 & 77.85& 64.19 & 55.76 &NA&NA  \\
Two-Phase \cite{lai2014facial}& HC  & 79.54 & 72.85& 74.46 &70.12& 58.57&50.98&NA&NA \\
LDTexP \cite{rivera2015local} & HC & 89.60 & 83.08 & 82.04& 78.70& 64.30 & 57.28&NA&NA  \\
RADAP \cite{mandal2019regional}& HC & 88.48 & 83.72 & 82.65& 78.57& 67.05 & NA &NA&NA \\
LDTerP \cite{ryu2017local}& HC & 85.89 &81.40& 80.15 & 78.11& 62.86& 55.76&NA&NA \\
VggNet 16 \cite{simonyan2014very} & CNN  & 91.31& 88.14& 85.14 &84.67& 73.09&59.32&138&528\\
VggNet19\cite{simonyan2014very} & CNN & 89.98 & 78.17 & 85.12& 85.12& 70.24 & \textbf{69.20}&144&549 \\
ResNet50 \cite{he2016deep}& CNN&88.05 & 85.51&76.36 &74.28  &61.59 &54.19&25&98  \\
MobileNet \cite{howard2017mobilenets}  & CNN & 68.59 &55.47&53.63  &40.84 & 56.36&45.33&4.25&16\\
MobileNetV2  \cite{sandler2018mobilenetv2}& CNN &70.84&59.18 & 41.43  &57.53& 48.80 &47.01&3.5&14\\
HiNet  \cite{verma2019hinet}& CNN &91.40&88.60 &87.80   &87.2 & NA &NA&1&5.3\\
\textit{RARITY} & \textit{HC} & \textit{90.90} & \textit{85.10}    & \textit{83.40} &  \textit{79.09}  & \textit{65.49}& \textit{59.53}          &NA&NA \\
\textbf{RARITY\_Net}  & \textbf{CNN} & \textbf{92.20} &\textbf{88.60} & \textbf{91.58} & \textbf{91.87}   & \textbf{75.35}  &\textbf{63.44}  &1.6&13.5\\      
\hline
\end{tabular}
\end{table*}
\subsubsection{HBEF Stream}
Inspired by the concept High-boost filtering (“A high boost filter is also known as a high frequency emphasis filter. A high-boost filter is used to retain some of the low-frequency components to aid in the interpretation of an image”) in image processing, we proposed a deep learning based approach: high boost edge filtering (HBEF) to capture the sharpened responses. The HBEF allows the network to extract and learn the fine details of expressive regions (eyebrow shapes, eye line, wrinkles, crinkles, lip lines, mouth shape, galebella texture etc.), which are responsible to define the emotion classes. Specifically, HBEF stream is responsible for extracting expression aware edge information by taking two different inputs: original image and RARITY encoded responses. Initially, Each RARITY encoded response is  convoluted  with  $3\times 3\times 8$  sized  kernel  followed  by  ReLU  in lateral  manner  to  embellish  the  appearance  features.  While, parallely  connected  two  different  scaled  convolutional  layers as $3\times 3\times 8$ and $7\times 7\times 8$ are employed on original input images to extract the high-level edge information of facial expressions. Furthermore, responses generated from RARITY images are subtracted from the each laterally connected responses of original images to capture the trivial edge information of expressions. Finally,  expressive  edge  features  are  boosted  by  subtracting trivial  edge  information  from  the  refined  high-level  features those are generated by applying $3\times 3\times 8$ and $7\times 7\times 8$ sized convolutional layers in a sequence of previously lateral layers as  shown  in  Fig. \ref{fig1}. The outcome of HBEF Stream is computed by using [Eq. \ref{eq1.1}- Eq. \ref{eq1.5}.]
\begin{multline}\label{eq1.1}
    HBEF\left(I(M,N)\right)=S\left(I(M,N)\right)-\mathbb{C}^{3\times 3,32}_{1}\\\left(L_{1}\right)||S\left(I(M,N)\right)-\mathbb{C}^{7\times 7,32}_{1}\left(L_{2}\right)
\end{multline}
\begin{multline}\label{eq1.2}
    S\left(I(M,N)\right)=[S_{0}\left(I(M,N)\right)||...||\\S_{7}\left(I(M,N)\right)]]
\end{multline}
where,
\begin{equation}\label{eq1.3}
    S_{b}\left(I(M,N)\right)=L_{\left \lfloor \frac{b}{4}  \right \rfloor}-\mathbb{C}^{3\times 3,4}_{2}\left(R_{l}\left(I(M,N)\right)\right)
\end{equation}
where, $\mathbb{C}^{z\times z,d}_{s}$ represents the convolution function with $z\times z$ filter size, depth $d$ and stride $s$. $R_{l}$ implies for the RARITY encoded outcomes. $l$ and $L$ are generated by using Eq. \ref{eq1.4} and [Eq. \ref{eq1.5}- Eq. \ref{eq1.5}].
\begin{equation}\label{eq1.4}
    l=\left[\{mod\left(b,4\right)\}\times 4\right]
\end{equation}
\begin{equation}\label{eq1.5}
    L_{1}=\mathbb{C}^{3\times 3,4}_{2}\left(I(M,N)\right)
\end{equation}
\begin{equation}\label{eq1.6}
     L_{2}=\mathbb{C}^{7\times 7,4}_{2}\left(I(M,N)\right)
\end{equation}
\begin{table*}[!t]
\centering
\caption{Performance Comparison on OULU-NIR Dataset for 6-class and 7-class Expressions.}
\label{tab2}
\begin{tabular}{lc|cccc|cccc}\hline 
\multirow{2}{*}{\textbf{Method} }& \multirow{2}{*}{\textbf{Type}} & \multicolumn{4}{c|}{\textbf{6-Class} }& \multicolumn{4}{c}{\textbf{7-Class}} \\ \cline{3-10} 
                  &                   &   \textbf{Weak}   &     \textbf{Dark}&  \textbf{Strong}   &   \textbf{Avg.}  &    \textbf{Weak}  &    \textbf{Dark} &   \textbf{Strong}   & \textbf{Avg.}     \\\hline
LBP \cite{shan2009facial}   &HC      & 63.26 & 64.09  & 67.09 & 64.81&61.84&63.63&65.71&63.72 \\
LDP \cite{jabid2010robust}  &HC      & 63.05 & 64.09  & 59.44 & 62.19&62.32&62.97&65.22&63.50 \\
LDN \cite{rivera2012local}  &HC      & 64.65 & 63.05  & 65.48 & 64.39&60.61&63.21&64.82&62.88 \\
Two-Phase \cite{lai2014facial}&HC   & 44.93 & 47.36  & 48.51 & 46.93&44.58&45.41&46.30&45.43 \\
LDTexP \cite{rivera2015local} &HC     & 62.36 & 61.11  & 67.77 & 63.74&62.14&61.13&66.72&63.33 \\
LDTerP \cite{ryu2017local}&HC     & 51.31 & 56.04  & 57.08 & 54.81&51.96&40.65&55.83&49.48 \\
RADAP \cite{mandal2019regional}&HC     & 64.65 & 65.69  & 65.76 & 65.30 &65.35&66.19&64.70&65.41 \\
ResNet50 \cite{he2016deep} &CNN   &  70.97     &  69.72      &68.19       &  69.62 &63.21&63.33&64.52&63.68    \\
MobileNet \cite{howard2017mobilenets}&CNN  & 54.79      &  59.58      & 53.48      &   55.95&44.34&52.23&56.13&53.36    \\
MobileNetV2 \cite{sandler2018mobilenetv2}&CNN & 62.08      & 60.48       & 62.10      & 61.55&52.91&52.90&54.28&53.36      \\
\textit{RARITY} &\textit{HC}     & \textit{63.47} & \textit{64.04}  & \textit{67.54} & \textit{65.01}& \textit{58.90}&\textit{63.86}&\textit{69.35}&\textit{64.03} \\
\textbf{RARITY\_Net} &\textbf{HC}& \textbf{75.55} &\textbf{76.24} &\textbf{78.12}& \textbf{76.63}&\textbf{72.26}&\textbf{65.59}&\textbf{76.01}&\textbf{71.28}\\\hline  
\end{tabular}
\end{table*}
\subsubsection{MSSEC Stream}
Multi-scale sophisticated edge cumulative (MSSEC) stream with  dense  connections is  proposed to overcome  the  issues  of  vanishing  gradient  in  deeper networks. MSSEC stream is designed to preserve  features related to affective facial appearance in high-level layers. The resultant features of MSSEC stream are generated by calculating [Eq. \ref{eq1.7}- \ref{eq1.9}].  
\begin{multline}\label{eq1.7}
     MSSEC\left(H\left (I(M,N)\right)\right)=\mathbb{C}^{3\times 3,96}_{2}\left(A_{2}\right)+\\\mathbb{C}^{5\times 5,96}_{2}\left(A_{2}\right)+\mathbb{C}^{7\times 7,96}_{2}\left(A_{2}\right)+\mathbb{C}^{1\times 1,96}_{2}\left(A_{2}\right)+\\\mathbb{C}^{1\times 1,96}_{4}\left(A_{1}\right)
\end{multline}
\begin{multline}
\label{eq1.8}
     A_{2}=\mathbb{C}^{3\times 3,64}_{2}\left(A_{1}\right)+\mathbb{C}^{5\times 5,64}_{2}\left(A_{1}\right)+\\\mathbb{C}^{7\times 7,64}_{2}\left(A_{1}\right)+\mathbb{C}^{1\times 1,64}_{2}\left(A_{1}\right)
\end{multline}
\begin{multline}
\label{eq1.9}
     A_{1}=\mathbb{C}^{3\times 3,32}_{2}\left(H\left (I(M,N)\right)\right)+\mathbb{C}^{5\times 5,32}_{2}\\\left(H\left (I(M,N)\right)\right)+\mathbb{C}^{7\times 7,32}_{2}\left(H\left (I(M,N)\right)\right)+\\H\left (I(M,N)\right)
\end{multline}

\subsubsection{RUCCF Stream}
The   RARITY uplift complementary context feature (RUCCF)  stream is   design   to   embedding   the   RARITY enriched  responses  as  a  complementary  context  features  to enhance  the  discriminative  feature  learning  of  the  model. Moreover,  RUCCF  is  constructed  in  such  a  way,  first  RARITY encoded responses are embellish by applying convolution layer of  $3\times 3\times 16$  size,  co-relative  to  all  four  responses.  Afterwards, important  features  are  summarized  in-terms  of  mean,  maximum  and  addition  followed  by  average  pooling  layers  as shown in Fig. \ref{fig1}. The outcome of RUCCF stream is computed by using [Eq. \ref{eq1.10}- \ref{eq1.12}]
\begin{equation}\label{eq1.10}
     RQ_{\varrho}=GAP\left[\varrho \{Q_{1}, Q_{2}, Q_{3}, Q_{4}\}\right]_{\varrho=1}^{\varrho=3}
\end{equation}
where, 
\begin{equation}\label{eq1.11}
     \varrho=\{1\rightarrow mean, 2\rightarrow max, 3\rightarrow add\}
\end{equation}
\begin{equation}\label{eq1.12}
     Q_{t}=\mathbb{C}^{3\times 3,16}_{2}\left(R_{t}\left (I(M,N)\right)\right)
\end{equation}
where, GAP and $R_{t}$ implies for the global average pooling and  RARITY encoded outcomes, respectively.\par
Moreover, the final result is computed by using [Eq. \ref{eq1.13}]\\
\begin{multline}\label{eq1.13}
     RARITYNet\left (I(M,N)\right)=FC^{E}[FC^{128}\\\left[FC^{128}\{\mathbb{P}\}\right]]
\end{multline}
\begin{multline}\label{eq1.14}
     \mathbb{P}=\{MSSEC\left(H\left (I(M,N)\right)\right)\}||RQ_{1}||\\RQ_{2}||RQ_{3}
\end{multline}
where, $FC^n$ and $||$ represents the fully connected layer with $n$ neurons and concatenation operation respectively.
\begin{table*}[!t]
\centering
\caption{Performance Comparison on OULU-VIL Dataset for 6-class and 7-class Expressions.}
\label{tab3}
\begin{tabular}{lc|cccc|cccc}\hline 
\multirow{2}{*}{\textbf{Method} }& \multirow{2}{*}{\textbf{Type}} & \multicolumn{4}{c|}{\textbf{6-Class} }& \multicolumn{4}{c}{\textbf{7-Class}} \\ \cline{3-10} 
                  &                   &   \textbf{Weak}   &     \textbf{Dark}&  \textbf{Strong}   &   \textbf{Avg.}  &    \textbf{Weak}  &    \textbf{Dark} &   \textbf{Strong}   & \textbf{Avg.}     \\\hline
LBP  \cite{shan2009facial}       & HC   & 57.70 & 55.90 & 75.34  & 62.98& 57.32 & 57.38 & 65.71  & 60.13 \\
LDP \cite{jabid2010robust}        & HC   & 56.45 & 46.90 & 72.43  & 58.60& 54.88 & 53.86 & 69.16  & 59.30  \\
LDN \cite{rivera2012local}         & HC   & 58.61 & 53.81 & 72.29  & 61.57& 59.28 & 53.86 & 70.89  & 61.34 \\
Two-Phase \cite{lai2014facial}  & HC   & 38.47 & 31.04 & 55.83  & 41.78& 39.22 & 28.09 & 53.98  & 40.43 \\
LDTexP \cite{rivera2015local}     & HC   & 55.62 & 42.36 & 72.29  & 56.75& 54.94 & 42.08 & 68.86  & 55.29 \\
LDTerP \cite{ryu2017local}      & HC   & 52.70 & 43.26 & 68.54  & 54.83& 48.75 & 40.65 & 64.53  & 51.31  \\
RADAP \cite{mandal2019regional}&HC     & 52.64 & 75.83  & 60.63 & 63.03& 52.97 & 74.11  & 61.07 & 62.72 \\
ResNet50 \cite{he2016deep}    & CNN  & 33.47      & 35.76      &       31.18 &  33.47& 53.03      &  38.38      &  63.69     &  51.70     \\
MobileNet \cite{howard2017mobilenets}  & CNN  & 25.38      &  35.83     & 21.38       & 27.53 &  39.94 & 27.87      & 39.82       & 35.87     \\
MobileNetV2 \cite{sandler2018mobilenetv2} & CNN  & 35.79      &  38.68     & 57.63       &   44.03 & 47.02     &  32.78     & 47.67       &  42.49    \\
\textit{RARITY}& \textit{HC} & \textit{57.84} & \textit{56.59} & \textit{75.76}  & \textit{63.40}& \textit{56.90} & \textit{56.13} & \textit{73.86}  & \textit{62.30} \\
\textbf{RARITY\_Net} & \textbf{CNN}  & \textbf{72.22}&\textbf{62.01}& \textbf{79.16}&\textbf{71.13}&\textbf{80.00}& \textbf{67.20}&\textbf{76.66}&\textbf{74.62}\\\hline    
\end{tabular}
\end{table*}

\section{Experimental Setups and Discussions}
In order to evaluate the performance of the proposed framework, we use four facial expression datasets: CK+, MUG, ISED and OULU-CASIA  datasets with various challenges ethnicity variations, multi-view, candid, low resolution, blurriness and illumination changes. We used Viola Jones algorithm to detect the face and normalized it to $120\times 120$. All experimental results are evaluated by adopting the leave-one-subject-out cross validation strategy. The proposed method follows the experimental setup (person independent (PI)) of RADAP \cite{mandal2019regional} for all datasets. Moreover, to validate the effectiveness of the proposed RARITY descriptor alone we have utilized the SVM classifier. To train proposed RARITYNet, we have initialized learning rate to $1e^-3$
with weight decay $2\times 1e^{-6}$and momentum 0.8. Stochastic gradient descent (SGD) is used for optimization. All
the experiments are performed on python 2.7. Furthermore, to mitigate the problem of overfitting, we have performed data augmentation to
create a enhance the datasamples in training. Specifically, each image is rotated between $[30^\circ,
-30^\circ]$ with the increment of $15^\circ$ and then all outcome
images are flipped horizontally. \par
In the literature, various data selection and experimental frameworks have been adopted for FER. Thus, it is difficult to make a fair comparison by directly using the published results. Moreover, most papers do not reveal the exact details of the person-dependent or independent N-fold strategy. Therefore, in this work, we have implemented existing descriptors and deep learning techniques and incorporated them into our experimental framework.

\begin{figure*}[!t]
\centering
\centerline{\includegraphics[width=\linewidth]{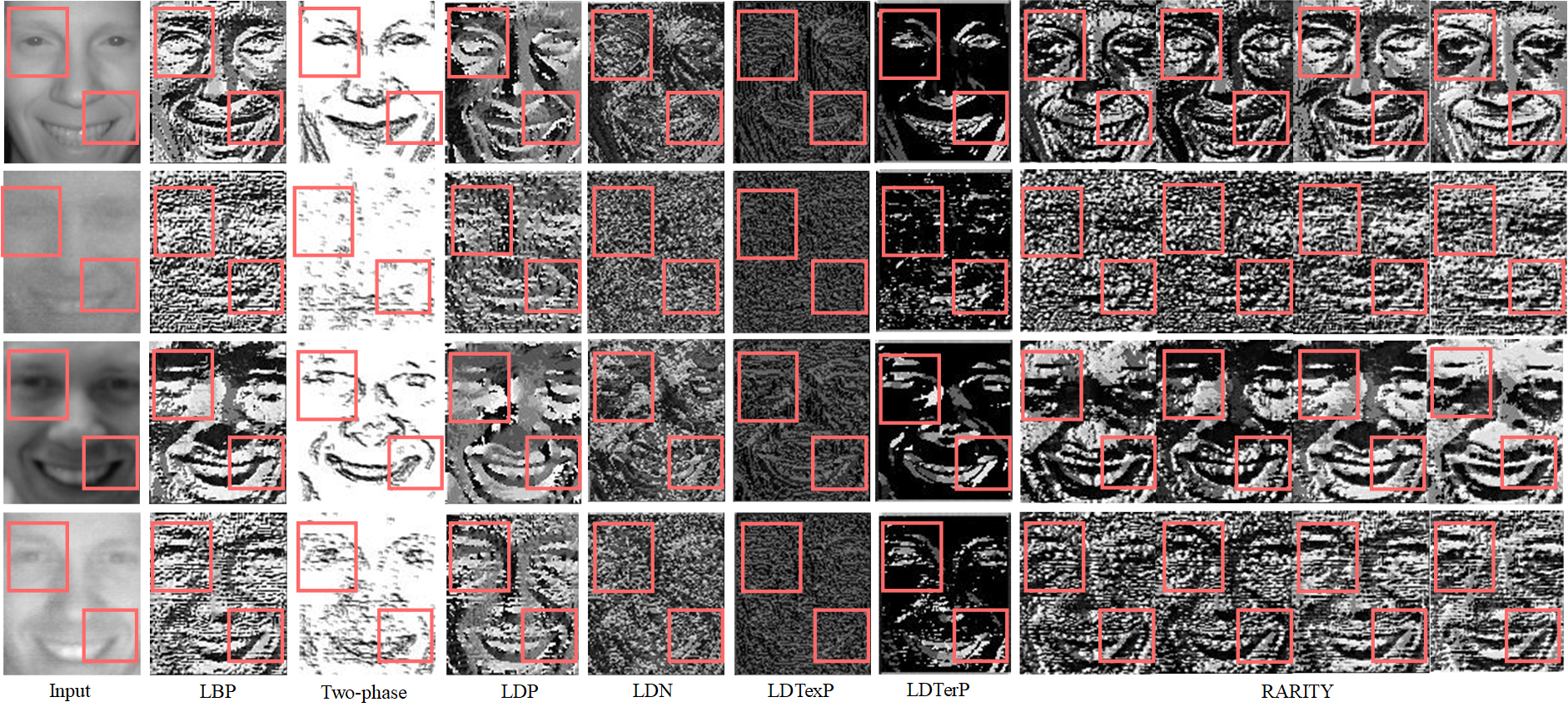}}
\caption{The input images and encoded response in various noise and illumination conditions by applying (a) LBP, (b) Two-phase, (c) LDP, (d) LDN, (e) LDTexP, (f) LDTerP and (g) RARITY. The qualitative difference between the RARITY and existing descriptors are highlighted in the rectangular regions. In 1st image, RARITY is able to extract the edge-spreads (in the nose, mouth and eyes regions) instead of extracting the extreme edges or the generic textures as detected by LDTexP, LBP, LDN and LDTerP. Moreover, in 2nd, 3rd and 4th image, the RARITY is more robust in preserving expressive features in comparison to other methods.}
\label{fig4}
\end{figure*}
\subsection{Quantitative Analysis}
In this section we presents the quantitative results of proposed descriptor: RARITY and CNN framework: RARITYNet, respectively.
\subsubsection{RARITY}
Recognition accuracy results over CK+, MUG, ISED
and OULU-VIS datasets for existing descriptors and
proposed RARITY for person independent setup are
tabulated in Table \ref{tab1}-\ref{tab3} respectively. The proposed Rarity descriptor outperforms LBP and LDTerP descriptors by 0.93\%, 1.14\% and 5.01\%, 3.7\% for 6- and 7-class problem over CK+ dataset. For MUG dataset, RARITY obtains 0.75\%, 2.93\% and 1.36\%, 0.39\% performance improvement over LBP and LDTexP for 6- and 7-class problem respectively. Similarly, in case of ISED, RARITY outperforms LBP and LDTerP by 2.2\%, 0.53\% and 2.63\%, 3.77\% for 6- and 7-class problems respectively. Moreover, for OULU-CASIA dataset, we have evaluated results for both NIR and VIS for all illumination conditions: weak, dark and strong by considering 6- and 7- classes as reported in Table \ref{tab2}-\ref{tab3}. Specifically, RARITY gains 0.2\%, 10.2\% and 0.31\%, 14.55\% more average recognition accuracy of all three lighting variations as compared to LBP and LDTerP for 6-and 7-class expressions over OULU-NIR. Similarly for OULU-VIL, the proposed model attains 0.42\%, 8.57\% and 2.17\%, 10.99\% more recognition performance as compared to LBP and LDTerP for 6-and 7-classes, respectively.
\subsubsection{RARITYNet}
Experimental results over CK+, MUG, ISED
and OULU-VIS datasets for existing and
proposed RARITYNet for person independent setup are
tabulated in Table \ref{tab1}-\ref{tab3} respectively. Specifically, for CK+, RARITYNet achieves 4.15\%, 3.09\% and 21.36\%, 29.18\%  more accuracy for 6- and 7-class as compared to the ResNet50 and MobileNetV2. In case of MUG dataset, RARITYNet obtains 15.22\%, 3.78\% and 17.59\%, 4.67\% performance improvement over ResNet50 and HiNet for 6- and 7-class problem respectively. The RARITYNet outperforms ResNet50 and MobileNetV2 by 26.55\%, 13.76\% and 16.43\%, 9.25\% for 6- and 7-class problems, respectively. The RARITY gains 7.01\%, 15.08\% and 7.6\%, 17.92\% more average recognition accuracy of all three lighting variations as compared to ResNet50 and MobileNetV2 for 6-and 7-class expressions over OULU-NIR. Similarly, the proposed framework outperformed 37.66\%, 27.1\% and 22.92\%, 32.13\%  ResNet50 and MobileNetV2 for 6-and 7-classes, respectively.

\subsection{Qualitative Analysis}
In this section we presents the qualitative results of proposed descriptor: RARITY and CNN framework: RARITYNet, respectively. The capability of RARITY towards the illumination invariance is depicted in Fig. \ref{fig4}. In Fig. \ref{fig4} (a, b, c, d), we have selected four images under different conditions (normal image, infrared image, noise and illumination changes). It can be seen from Fig. \ref{fig4} that; the proposed descriptor is able to identify the relevant expression features even with the noise and illumination variations. The qualitative visualization of state-of-the-art CNN models: ResNet-50, MobileNet, MobileNetV2 and proposed RARITYNet is demonstrated in Fig. \ref{fig6}. Fig. \ref{fig6} includes the mean neurons visualization for two test expressions: surprise and sad of the two different datasets: MUG and ISED, respectively. From Fig. \ref{fig6}, it visibly clear that the proposed RARITYNet neurons are able to capture more generalized and discriminative features as compared to other very popular CNN architectures. For example, in surprise: mouth, eyes and eyebrow shapes regions and in sad: eyes, lower chin part, nose and glabella regions are play significant role to make a decision. The proposed RARITYNet is able to surpass the other regions and captures the most prominent regions. 
 
\begin{figure}[!t]
\centering
\centerline{\includegraphics[width=\linewidth]{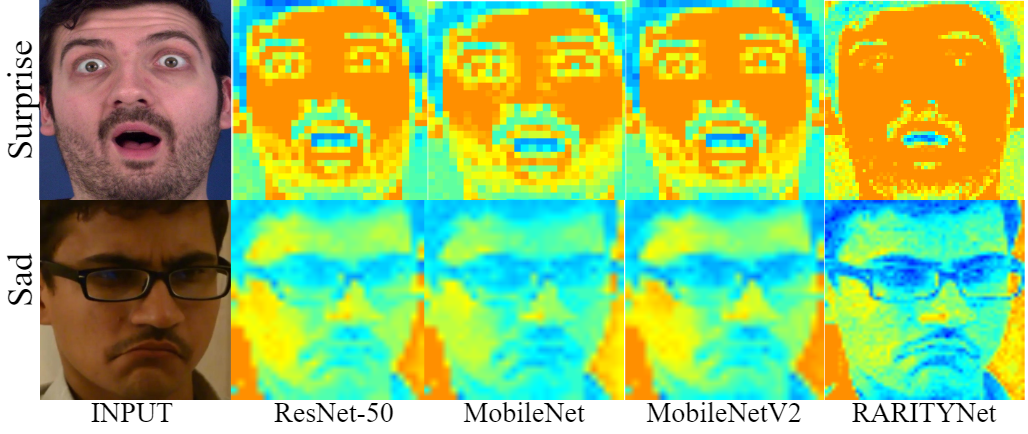}}
\caption{Visualization of neurons computed from two distinctive emotion classes a) Sad and b) Surprise from MUG and ISED, respectively.}
\label{fig6}
\end{figure}

\section{Conclusion}
In this paper, we proposed a RARITY guided affcetive emotion learning framework for facial expression recognition. The RARITYNet is designed to capture the facial appearance feature by learning shallow (RARITY) and deep (AffEmoNet) features. 
The RARITY descriptors is proposed to extracts the inter radial topological information in the local region. RARITY computes the edge spread features in the local region to address the intra-class variation in facial expression classes. Moreover, it establishes the pixel triplet relations to cohesively capture the changes in expression images. Furthermore, to enhance the robustness of the proposed descriptor, bi-directional relations between inter-radial rings are identified. These properties increase the robustness of the proposed method by identifying the edge spread in the local region, highlighting the salient features and suppressing the noise elements. Moreover, the RARITY encoded features are processed through the AffEmoNet to learn the macro and micro level features from the faces. The guidance of RARITY features reduced the computation cost of the AffEmoNet as well as allows to learn significant features of facial structures in various challenging scenarios. The experimental results show that the proposed method obtains better recognition rates as compare to other state-of-art methods over 4 benchmark facial expression datasets. 

\bibliographystyle{IEEEtran}
\bibliography{Rarity}

\newpage
\begin{table}[bp]
\centering
\caption{The Number of Images used in the experiments for  Person Independent (Leave One subject Out) Frameworks.}
\begin{tabular}{ccccc}\hline
\textbf{Emotion}  & \textbf{CK+}  & \textbf{MUG}  & \textbf{ISED} & \textbf{OULU-VIS/NIR} \\ \hline \hline
\textbf{Anger}    & 152  & 220  & -    & 240/240      \\
\textbf{Disgust}  & 190  & 220  & 234  & 240/240      \\
\textbf{Fear}     & 144  & 220  & -    & 240/240      \\
\textbf{Happy}    & 247  & 220  & 294  & 240/240      \\
\textbf{Sad}      & 181  & 220  & 174  & 240/240      \\
\textbf{Surprise} & 246  & 220  & 189  & 240/240      \\
\textbf{Neutral}  & 305  & 220  & 295  & 240/240      \\
\textbf{Total}    & 1465 & 1540 & 1186 & 1680/1680   \\ \hline
\end{tabular}\label{tab1}
\end{table}
\section{Supplementary Document}
\chapterinitial{The introduction}
This supplementary represents the detailed statistics of data samples, qualitative analysis of RARITY as compared to state-of-the-art descriptors. The detailed description of all datasets is tabulated in \ref{tab1}. \par
The proposed descriptor utilizes the inter radial transitional information between the rings at multiple radius. This leads to identification of the edge spreads in the local neighborhood. These edge spreads capture the edges using higher distanced neighbors and therefore are more robust to intra-class variations among facial expressions in an emotion class. 

\begin{figure}[!t]
\centering
\centerline{\includegraphics[width=\linewidth]{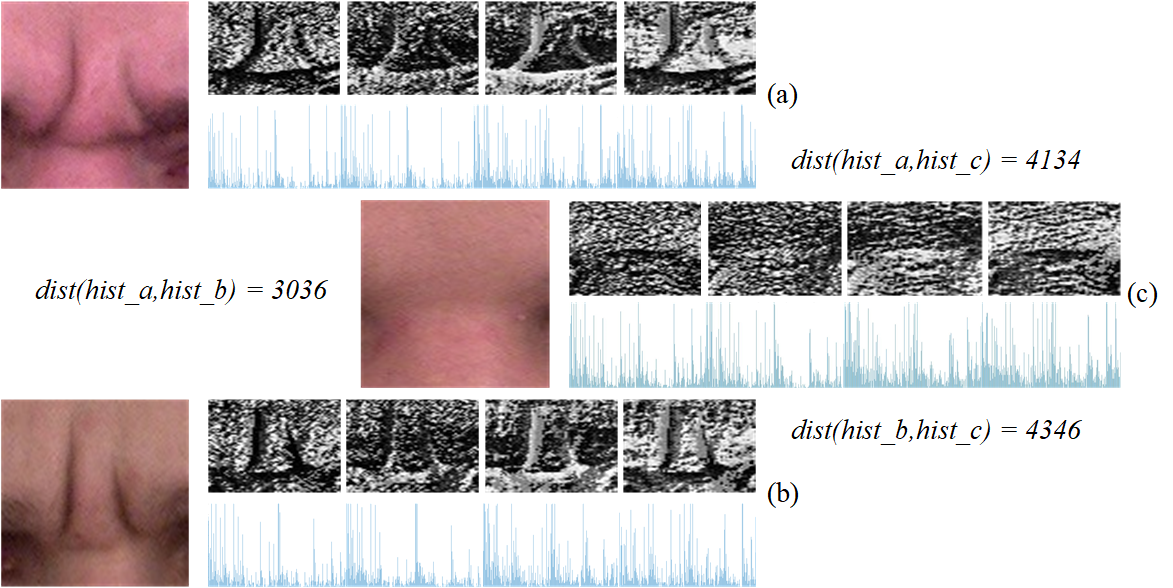}}
\caption{The active facial patches, their RARITY feature maps and corresponding histograms. The first two patches (a), (b) are selected from two different subjects belonging to anger emotion class. The last patch (c) is selected from a subject belonging to neutral emotion class. The Manhattan distance between the feature vectors of (a) and (b) is very less as compared to the distance between (a), (c) and (b), (c).}
\label{fig3}
\end{figure}

To demonstrate this property, we have selected two different facial images from the ‘anger’ emotion class. Although, these facial images belong to the same emotion category, they all differ from each other with respect to the intensity variations at different locations. From each image, an active patch (i.e. region with salient expression changes) is selected for further analysis. In Fig. \ref{fig3}, we have shown the RARITY response and the corresponding histograms for these active patches. From Fig. \ref{fig3} (a, b), we can see that, the proposed method is able to categorize these patches to the same class even though there are minor variations in the edge positions in the original images. Also, we have depicted the RARITY feature maps of a different emotion category ‘neutral’ in Fig. \ref{fig3} (c). Furthermore, along with the visual representation, we have calculated the similarities between the intra-class images using Manhattan distance. We have also computed the Manhattan distance between the inter-class images. These distance measures demonstrate the robustness of the proposed RARITY to intra-class variations as well as the ability to distinguish between inter-class expressions. Since, the RARITY patterns are encoded using pixel triplet relationship between two rings of different radius. This enhances the discriminability by highlighting the salient features (maximal and minimal disparities) and suppressing the noise elements in an image. \par

\begin{figure}[!t]
\centering
\centerline{\includegraphics[width=\linewidth]{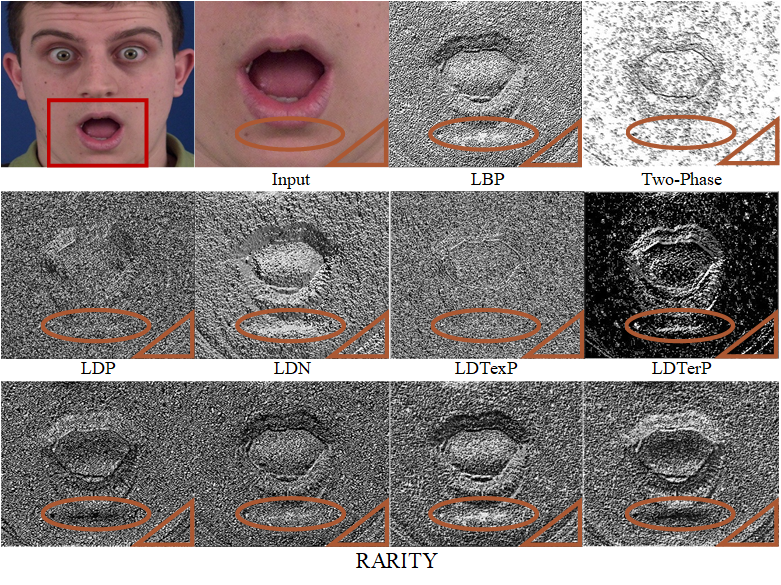}}
\caption{The feature maps of an expression image of surprise emotion class. The RARITY is more robustly detecting the crease below the lower lip and jaw-line features as compared to other feature descriptors.}
\label{fig5}
\end{figure}

The authors in LDN, LDTerP and LDTexP have used 8 compass masks (Robinson \& Kirsch) to compute the edge responses. The coded patterns were generated by analyzing the salient directional index locations in the local neighborhood. However, the proposed descriptor is able to capture the salient change information by encoding the bidirectional patterns using pixel triplets as shown in Fig. \ref{fig5}. These patterns are generated by considering inter radial relations. The most recent state-of-the-art descriptor LDTerP focuses mainly on the extreme edge variations in a region.
This may lead to information loss in the facial region thereby degrading the performance as shown in Fig. \ref{fig6}. Whereas, the inter radial and bidirectional properties of RARITY enhance the capability to robustly identify the edge spread. From the Fig. \ref{fig5} and \ref{fig6}, it is also clear that proposed RARITY descriptor outperformed the state-of-the-art descriptors.

\begin{figure}[!t]
    \centering
    \centerline{\includegraphics[width=\linewidth]{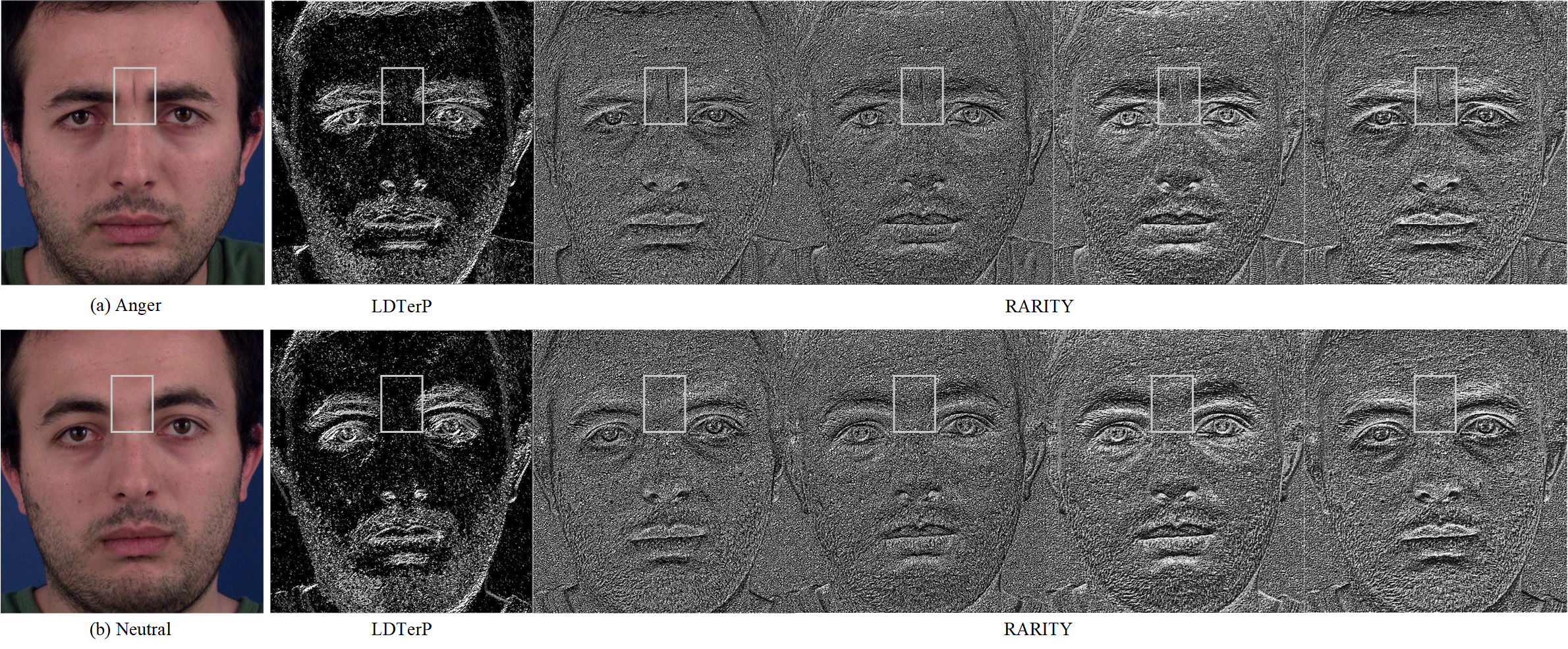}}
    \caption{The qualitative comparison between the feature maps generated by LDTerP and RARIRTY descriptor. The region of interest shows that RARITY is able to detect the furrow lines more accurately as compared to LDTerP.}
    \label{fig6}
\end{figure}
\end{document}